\documentclass{uai2024}

\usepackage[american]{babel}

\usepackage{natbib} 
\usepackage{times}
\usepackage{subfig}
\usepackage{hyperref} 
\usepackage{url}            
\usepackage{booktabs}
\usepackage{amsfonts}
\usepackage{nicefrac}
\usepackage{microtype}
\usepackage{graphicx}
\usepackage{tikz}
\usepackage{multirow}
\usepackage{comment}
\usepackage{enumitem}
\usepackage{xcolor}
\usepackage{framed}
\usepackage{algorithm}
\usepackage{algpseudocode}
\usepackage{mathtools}
\usepackage[capitalize,noabbrev]{cleveref}

\author[1]{\href{mailto:<amin.soleimaniabyaneh@mail.mcgill.ca>?Subject=Federated Causal Discovery Paper}{Amin Abyaneh}{}}
\author[2]{Nino Scherrer}
\author[3]{Patrick Schwab}
\author[4]{Stefan Bauer}
\author[5]{Bernhard Sch\"olkopf}
\author[3, 5, 2]{Arash Mehrjou}

\affil[1]{%
    Electrical and Computer Engineering Dept., 
    McGill University,
    Montréal, Canada
}
\affil[2]{%
    ETH Z\"urich, Z\"urich, Switzerland
}
\affil[3]{%
    GlaxoSmithKline, Artificial Intelligence and Machine Learning, Zug, Switzerland
  }
\affil[4]{%
     TU Munich \& Helmholtz AI, Munich, Germany
  }
\affil[5]{%
    Max Planck Institute for Intelligent Systems,  T\"ubingen, Germany
  }

\newcommand{\ours}{FedCDI}
\newcommand{\STATE}{\State}
\definecolor{shadecolor}{RGB}{250, 250, 100}

\newtheorem{theorem}{Theorem}[section]
\newtheorem{proposition}[theorem]{Proposition}

\newtheorem{definition}[theorem]{Definition}
\newtheorem{assumption}[theorem]{Assumption}

\newtheorem{example}{Example}[theorem]

\title{Federated Causal Discovery from Interventions}

\begin{document}

\maketitle

\begin{abstract}  
\emph{Causal discovery} serves a pivotal role in mitigating model uncertainty through recovering the underlying causal mechanisms among variables. In many practical domains, such as healthcare, access to the data gathered by individual entities is limited, primarily for privacy and regulatory constraints. However, the majority of existing causal discovery methods require the data to be available in a centralized location. In response, researchers have introduced \emph{federated} causal discovery. While previous federated methods consider distributed observational data, the integration of \emph{interventional data} remains largely unexplored. We propose \ours{}~\footnote{\href{https://github.com/aminabyaneh/fed-cdi}{github.com/aminabyaneh/fed-cdi}}, a federated framework for inferring causal structures from distributed data containing interventional samples. In line with the federated learning framework, \ours{} improves privacy by exchanging belief updates rather than raw samples. Additionally, it introduces a novel intervention-aware method for aggregating individual updates. We analyze scenarios with shared or disjoint intervened covariates, and mitigate the adverse effects of interventional data heterogeneity. Performance and scalability of \ours{} is rigorously tested across a variety of synthetic and real-world graphs.
\end{abstract}


\section{Introduction}
\label{sec:introduction}

The fundamental and challenging problem of discovering cause-effect relationships has long captivated the attention of statisticians~\citep{pearl2009causality}. Recovering causal structures from data provides valuable insights for decision-making under uncertainty, as precise causal models of covariates empower exploration of potential outcomes of interventions and hence, mitigate the unpredictability caused by external factors~\citep{bhattacharya2020causal-uncertainty, vowels2021d}. \looseness-1

Conventional causal discovery approaches mainly rely on centralized data~\citep{wang2017permutation, zheng2018dags, ke2020dependency}. However, when dealing with sensitive information, such as patients' medical records, preserving privacy is essential and protected by strict regulations. As a result, such sensitive data remains \emph{confidential} and \emph{distributed} across multiple entities. Therefore, researchers have turned to federated causal discovery (FCD)~\citep{li2020federated} to discover causal structures in decentralized settings. \looseness-1

Previous FCD efforts are either limited to data generated from a linear underlying causal mechanism~\citep{ye2022distributed, huang2023fed-pc, mian2023regret-based-fcd} or restricted to homogeneous data~\citep{ye2022distributed, ng2022fed-bayesian-nets}. Most notably, these methods are \emph{merely oriented towards discovery from observational data}, gathered passively without perturbing the system. This exclusive reliance on observational samples confines the identified structures to Markov equivalence classes~\citep{yang2018characterizing, spirtes2016causal} and significantly elevates model uncertainty~\citep{jesson2020causality-uncertainty-aware}. Utilizing interventional data, collected by applying targeted perturbations on the system~\citep{hyttinen2013experimentdesign, addanki2020efficientintervention}, alleviates these limitations through enhancing identifiability~\citep{wang2017permutation, brouillard2020differentiable, ke2020dependency, tigas2022interventions-how}. 

\begin{figure}
    \centering
    \includegraphics[width=0.9\linewidth]{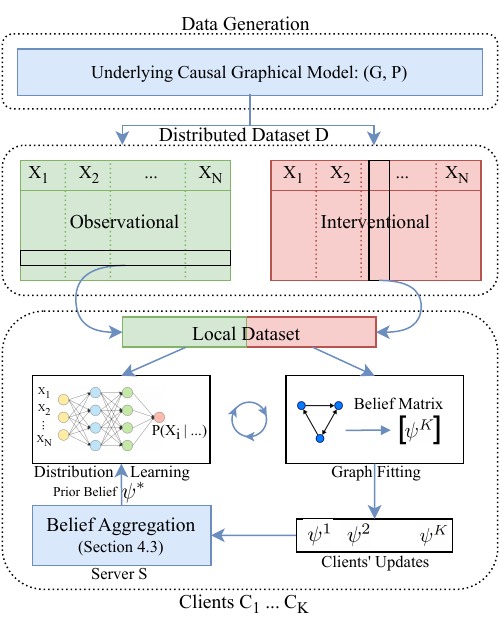}
    \caption{Overview of \ours{}, including the distributed dataset and local learning process. We only depict a single client, as the rest perform similar operations.}
    \label{fig:overview_fedcd}
\end{figure}

To address the absence of interventional data in federated settings, we present \ours{}: a federated causal discovery method relying on interventions. \ours{} learns a global causal structure from a federation of clients (entities), without direct access to their local data. To this end, clients adopt a neural causal discovery method, such as \citet{lippe2021efficient}, to acquire a belief about the causal structure of their local data. These local findings are then aggregated on a server while conforming to the following two criteria: (1) clients exert a more substantial influence on the aggregated belief based on the quality of their interventional samples, and (2) data samples are not exchanged among clients or transmitted to the server. Instead, adhering to the federated learning paradigm, only belief updates are transmitted to the server during the aggregation process. An overview of \ours{} is presented in \Cref{fig:overview_fedcd}.

Conforming to the mentioned criteria, \ours{} effectively aggregates clients' updates based on the reliability of their local data. Owing to our novel knowledge aggregation method, clients' contributions are rated by the extent of their access to interventional samples and the location of interventions in the causal structure. Therefore, a client provides a stronger contribution to the existence of a cause-effect relationship in the vicinity of its intervened covariates. Moreover, our approach poses few restrictive assumptions about the data generation mechanism, and endures a certain level of heterogeneity in interventional data. We empirically demonstrate that \ours{} performs similar to the centralized approaches, and outperforms FCD methods, when applied to data from synthetic and real-world causal structures.

\section{Related Work}
\label{sec:related_work}

Attempts to discover causal structures from distributed data precede the rise of federated learning. However, such distributed methods typically operate for a single round with no iterative improvement, and the majority decides the discovered DAG. For example, \citet{gou2007learning} propose a two-step approach: obtaining a local Bayesian network, and utilizing conditional independence tests to consolidate local updates. Moreover, \citet{na2010distributed} develop a voting-based approach where the final structure is determined by most frequent patterns among local DAGs. 

More recently, there has been a surge in FCD methods. FedDAG~\citep{gao2021feddag} is one of the first where at each round, clients learn an adjacency matrix to estimate the DAG and neural representation to approximate causal mechanisms. \citet{ng2022fed-bayesian-nets} employ ADMM~\citep{boyd2011distributed} and apply a centralized optimization proposed by \citet{zheng2018dags}, to a federated setting. DARLS~\citep{ye2022distributed} is another method that counts on generalized linear models and distributed optimization to find an optimal DAG without violating clients' privacy. Authors of PERI~\citep{mian2023regret} avoid the exchange of model parameters through regret-based aggregation to address privacy violations in previous methods, and \citet{wang2023towards} extend conditional independence test to enable FCD with data heterogeneity. \looseness-1 

Nevertheless, existing FCD methods face serious limitations, such as confinement to linear causal mechanisms or restricted to homogeneous data~\citep{ng2022fed-bayesian-nets, mian2023regret-based-fcd, ye2022distributed}. Notably, to the best of our knowledge, none of these methods, harness the valuable insights offered by interventional data. This underscores a \emph{critical gap} between current FCD approaches, and the literature of DAG discovery based on interventions~\citep{hauser2015jointly,wang2017permutation,ke2020dependency, brouillard2020differentiable}. Although interventional data is more expensive to collect, researchers demonstrate that efficient utilization of such data significantly alleviates the identifiability challenge~\citep{tigas2022interventions-how}.

Lastly, in this work, we build on DAG discovery methods based on continuous optimization. Unlike score-based methods~\citep{hauser2012characterization}, this class of algorithms avoids the combinatorial greedy search over DAGs through gradient-based optimization. DAGs with NO TEARS \citep{zheng2018dags, yu2019dag} is considered the first to reformulate the greedy search as a continuous optimization problem~\citep{vowels2021d}. SDI~\citep{ke2020dependency}, DCDI~\citep{brouillard2020differentiable}, and ENCO~\citep{lippe2021efficient} are introduced more recently as neural DAG discovery techniques to leverage interventional samples on top of regular observational data. Among the neural frameworks, ENCO~\citep{lippe2021efficient} presents a reliable and efficient DAG discovery method which formulates the graph search as optimization of independent edge likelihoods, while remaining scalable to large feature spaces.

\section{Background and Problem Setup}
\label{sec:problem_statement}

\emph{Causal graphical models} (CGM) provide mathematical abstraction to quantitatively describe causal relations between random variables. A CGM is a pair $(G, P)$, where $G = (X, E)$ is a \emph{directed acyclic graph} (DAG), and $P$ is a distribution faithful to $G$ \citep{spirtes2010introduction}. Within $G$, each node, $X_i \in X \; \forall i \in \{1, ..., N\}$, represents a random variable, and a directed edge, $X_i \rightarrow X_j$, denotes the existence of a cause-effect relationship from $X_i$ to $X_j$ \citep{pearl2009causality}. We denote the adjacency matrix of $G$ by $adj(G)_{N \times N}$. 
\begin{assumption} (Causal Sufficiency)
    We assume causal sufficiency of the CGM, i.e., all common causes of variables are included and observable.
\end{assumption}
An \emph{intervention} on a random variable $X$ is an applied perturbation on the system, such that it overrides the natural values of $X$ that affects its probability distribution and (possibly) alters the DAG. We call the gathered data from an unperturbed system \emph{observational}, $D_{\mathcal{O}}$, and we use the term \emph{interventional}, $D_{\mathcal{I}}$, when the data is gathered from the perturbed system. The collection of these two datasets is denoted by $D := \{D_{\mathcal{O}}, \; D_{\mathcal{I}}\}$, that is a centralized dataset generated by unperturbed and perturbed CGM described by $(G, P)$.
\begin{assumption} (Perfect Interventions). Interventions in $D_{\mathcal{I}}$ are perfect, that is,  $p_{\mathcal{I}}(X_i\mid\text{Pa}(X_i)) = p_{\mathcal{I}}(X_i)$, where $p_{\mathcal{I}}$ is the distribution of intervened $X_i$, and $\text{Pa}(X)$ denotes the parent variables of $X$ within the DAG.
\end{assumption}
\begin{assumption} (Stochastic and Known Interventions).
    Interventions are not designed to set the variable to a fixed value, and the intervened variable is known.
\end{assumption}
Our proposed federated setup consists of a central node acting as a server and $K$ other nodes as clients. We denote the server with $S$ and the clients with, $C_k, \; \forall k \in \{1, \ldots, K\}$. Each client is an independent processing unit, and can only communicate with the server. The centralized dataset $D$ is distributed among the clients, where each client $C_k$ has access to a (not necessarily disjoint) subset of $D$, called $D^{k} := \{D_{\mathcal{O}}^k, \; D_{\mathcal{I}}^k\}$. To utilize the entire dataset for structure discovery, we assume these subsets to span the entire $D$, such that $D = \bigcup^K_{k = 1} D^{k}$. 

Any dataset can be distributed either horizontally or vertically. Horizontal distribution leads to clients having access to the entire set of features, but only a fraction of samples. A vertical split, however, provides access to only a subset of features in the original centralized data. We consider a horizontal (homogeneous) split for $D_{\mathcal{O}}$ and vertical (heterogeneous) distribution of $D_{\mathcal{I}}$. Therefore, we assume that clients are aware of all the dataset features, but might not have access to interventional data corresponding to each node, $X_i$. The accessible set of intervened variables for each client, $C_k$, is represented by $X^{k}_{\mathcal{I}} \subset X$.

\begin{assumption} (Knowledge of Intervened Covariates).
    The information about $X^{k}_{\mathcal{I}}$, that is, the intervened covariates for $C_k$, is known to the server $S$.
\end{assumption}

Note that unlike existing methods, we do not pose any restrictions on linearity or noise distribution of the CGM.

\paragraph{Problem statement.} Within the described federated setting, we focus on recovering the underlying causal DAG, $G$, rather than the probability distribution, $P$, while enforcing a strict restriction on direct exchange of $D^{k}$ samples either between two arbitrary clients, $C_i$ and $C_j$, or to the server, $S$. \looseness-1

\section{Federated Causal Discovery From Interventions}
\label{sec:methodology}
Following the federated learning paradigm, clients in \ours{} collaborate through a two-phase iterative process:


\begin{enumerate}
    \item Local discovery (\Cref{sec:local_learning_stage}): Each client applies a local causal discovery method (LCDM) to yield a \emph{prior belief} about $G$ while incorporating an existing belief. Therefore, clients, $C_k$, securely share their belief with the server and not data samples or model parameters.

    \item Global aggregation (\Cref{sec:belief_aggregation_unbalanced_interventions}): The server employs an aggregation method to find the updated belief from a collection of clients' communicated beliefs. Then, the server broadcasts the \emph{aggregated belief} back to all clients, which they utilize to guide their local optimization in the next round.
\end{enumerate}

We start with the first step in \Cref{sec:local_learning_stage} by discussing the nature of the prior belief and LCDMs. In \Cref{sec:belief_aggregation_identical_interventions}, we assume a horizontal distribution of $D_{\mathcal{I}}$, that is, every client has access to the same set of interventions. Yet, assuming identical intervened covariates is a naive assumption for real-world scenarios, where clients oftentimes hold interventional data on various covariates. Hence, in \Cref{sec:belief_aggregation_unbalanced_interventions}, we propose a proximity-based belief aggregation applicable to clients with non-identical sets of intervened variables, i.e., vertical distribution of $D_\mathcal{I}$. This simulates a level of heterogeneity between clients in interventional data, since the underlying data distributions are different across clients.  

\subsection{Local discovery of prior beliefs}
\label{sec:local_learning_stage}
The goal in this stage is for each client to infer a belief about the global DAG. To this end, clients apply an LCDM to their local datasets, $D^k$. We design the LCDM to produce and later incorporate prior knowledge about the structure, and favor this knowledge when uncertain about the existence of any $X_i \rightarrow X_j \in E$. We define clients' beliefs with matrices of the same dimensions as $adj(G)$.
\begin{definition} 
\label{def:prior_beliefs}
\textbf{(Belief Matrix)} Let $\psi \in \mathbb{R}^{N \times N}$ be the matrix with elements $\psi_{ij}$, where $\psi_{ij}$ is the parameter of an independent Bernoulli distribution, i.e., \text{Ber}($\psi_{ij}$)$ \; \forall i, j$. We call $\psi$ the belief matrix, and each element, $\psi_{ij}$, corresponds to the existence probability of $X_i \to X_j$, i.e., $P(X_i \to X_j) \sim \text{Ber}(\mathcal{B}_{ij})$. The matrix $\psi$ yields an adjacency matrix through an element-wise binary step function.
\end{definition}

To learn $\psi$, we opt for neural LCDMs that learn similar belief matrices through continuous-optimization. For instance, ENCO \citep{lippe2021efficient} uses $\gamma, \; \theta \in \mathbb{R}^{N \times N}$ matrices to represent the existence and orientation of edges in a graph, respectively. This formulation is not unique to ENCO and other methods, such as SDI \citep{ke2020dependency} and DCDI \citep{brouillard2020differentiable} use similar belief matrices. We predominantly view ENCO as the preferred choice for clients' LCDM due to its efficient implementation and scalability.

The LCDM needs to consider the updated belief from the server into its local optimization. We achieve this by directly initializing the LCDM belief matrices, $(\gamma, \; \theta)$ in the case of ENCO, by sampling from Bernoulli distributions with $\psi$ parameters, and enforcing this knowledge as a Lagrangian on LCDM's internal loss. To implement this, we take note that ENCO alternates between \emph{distribution fitting} and \emph{graph fitting} stages. Distribution fitting trains a neural network $f_{\phi_i}$ to model $X_i$'s conditional data distribution, $p(X_i\mid\mathbf{X_{-i}})$, on observational data. Then, the graph fitting stage learns $\gamma$ and $\theta$ matrices by minimizing:
\begin{multline}
\label{eq:enco_graph_fitting_objective}
\mathcal{L} = \mathbb{E}_{\hat{I}\sim p_{I(I)}} \mathbb{E}_{\tilde{p}_{\hat{I}}(X)} \mathbb{E}_{p_{\gamma,\theta}(\zeta)}\left[\sum_{i=1}^{N}\mathcal{L}_{\zeta}(X_i)\right] + \\ \lambda_{sparse} \sum_{i=1}^{N}\sum_{j=1}^{N}\sigma(\gamma_{ij})\cdot\sigma(\theta_{ij}),
\end{multline}
where $p_{I}(I)$ and $\tilde{p}_{\hat{I}}(X)$ represent interventional data distribution and $p_{\gamma, \theta}(\zeta)$ is the distribution over adjacency matrices with $\zeta_{ij}\sim\text{Ber}(\sigma(\gamma_{ij})\sigma(\theta_{ij}))$. Note that $\sigma$ is the sigmoid function. Moreover, $\mathcal{L}_{\zeta}(X_i)=-\log f_{\phi_i}(X_i;\zeta_{\cdot,i}\odot X_{-i})$, that is the negative log-likelihood estimate of the variable $X_i$ with the distribution learned from observational data. The last term is for sparsity regularization~\footnote{Details of the optimization process and its tractability are provided in \citet{lippe2021efficient}.}. We integrate an extra differentiable loss term to consider the prior belief in \Cref{eq:enco_graph_fitting_objective}, yielding $\tilde{\mathcal{L}} = \mathcal{L} + \lambda_{prior} \mathcal{L}_{\psi}$, and define the belief loss as,
\begin{gather}    
\label{eq:enco_graph_fitting_with_prior}
     \mathcal{L}_{\psi} = -\mathbb{E}_{i, j} \left[ \zeta_{ij} \log(\psi_{ij}) + (1 - \zeta_{ij}) \log(1 - \psi_{ij}) \right],
\end{gather}
where the log loss measures the agreement between the observed values in $\zeta$ and the probabilities represented by $\psi$. Lower values of the log loss indicate a better agreement between $\zeta$ and $\psi$, encouraging the optimizer to find solutions that align with the prior belief. Hence, the optimization process of LCDM is guided by the belief, especially when facing uncertainty about the existence of an arbitrary edge, $X_i \rightarrow X_j$. Such uncertainties may arise when $D^k_{\mathcal{I}}$ lacks intervened covariates influencing the edge, $X_i \rightarrow X_j$.  

\newcommand\NormalComment[1]{{\footnotesize \color{cyan} /* {#1} */}}
\newcommand\InlineComment[1]{{\footnotesize \color{cyan} \Comment{{#1}} }}

\begin{algorithm}[ht]
\caption{Federated causal discovery from observational and interventional data. The algorithm for proximity-based aggregation method is provided in \Cref{alg:federated_setup_aggregation}.}
\label{alg:federated_setup_overview}
\begin{algorithmic}
\STATE \textbf{Data:} $D := \{D_{\mathcal{O}}, \;\; D_{\mathcal{I}}\}$;  $T := \text{n\_rounds} \in \mathbb{N}$ 
\STATE \textbf{Return:} $\tilde{G}:= \text{discovered causal DAG}$ 

\STATE \textbf{for} {$T \;\text{iterations}$} \textbf{or} convergence \textbf{do} \InlineComment{Federated rounds}
\STATE \quad \textbf{for} $k = 1$ \textbf{to} $K$ \textbf{do}
\STATE \quad \quad $\psi^k \gets \text{local\_causal\_discovery\_method}(D^{k}, \;\psi)$ 

\InlineComment{Parallel local learning (section 4.1)} 
\STATE \quad \textbf{end for}
\STATE \quad $\psi \gets \text{proximity\_based\_aggregation}(\psi, \;\{\psi^1, \ldots, \psi^k\})$ 

\InlineComment{Model aggregation stage (section 4.3)}
\STATE \textbf{end for}
\STATE $adj(\tilde{G}) \gets \text{binary\_step}(\psi^*,\; threshold=0.5)$

\end{algorithmic}
\end{algorithm}

\subsection{Naive belief aggregation}
\label{sec:belief_aggregation_identical_interventions}
In a naive scenario, we assume $X^{k}_{\mathcal{I}} = X, \; \forall k \in \{1, \ldots, K\}$. Let $\psi^k$ be the prior belief obtained by each client, after transmission of $\psi^k$ to the server, the aggregated belief is calculated by a weighted average over $\psi^k$ matrices through,
\begin{gather}
    \psi^* = \sum_{k} w_k \psi^k, \; w_k = \frac{size(D^k)}{size(D)}, \; \forall k \in \{1, \ldots, K\},
\end{gather}
where $\psi_{ij}^*$ is the aggregated belief. Because $w_1+ w_2+\ldots, w_K=1$, each element of the weighted sum $\psi_{ij}$ represents the parameter of a Bernoulli distribution. However, without access to same intervened covariates across clients' interventional dataset, this method fails to take client's expertise into account. Intuitively, a client with access to direct interventions on $X_i$ is more confident about the edge probability of $X_i \rightarrow X_j$ than a client without such knowledge. Nonetheless, this approach can serve as a naive baseline designed for homogeneous $D^k$ distributions, filling the void of FCD methods that harness interventional data.

\subsection{Proximity-based belief aggregation}
\label{sec:belief_aggregation_unbalanced_interventions}
We know that $D^k$, contains interventions on the set of variables indicated by $X^{\mathcal{I}}_{C_k}$. After applying LCDM to the local data, $\psi^{k}_{ij}$ is the posterior probability of $X_i \rightarrow X_j$'s existence, calculated by the client $C_k$. The challenge is to come up with an aggregation method for posteriors, $\psi^{k}_{ij} \; \forall k \in \{1, \ldots, K\}$, while considering the clients' knowledge of applied interventions. 

\begin{proposition}
Consider $X_i, X_j \in X$ to be two arbitrary variables in $G$. Then, for a client $C_k$ admitting an intervened variable, $X_s \in X^{k}_{\mathcal{I}}$, the reliability of $\psi^k_{ij}$, denoted by $r^k_{ij} \in [0, 1]$, is higher if $X_i \to X_j$ belongs to a path in $G$ descending from $X_s$. 
\end{proposition}

\noindent \textit{Sketch of the proof.} Note that if the described path exists, then every variable $X_i$ in the described path is a descendant of $X_s$. The proof is straightforward through causal factorization \citep{spirtes2010introduction}, 
\begin{gather}
\label{eq:causal_factorization}
P(X) = \prod_{i=1}^N P(X_i \mid \text{Pa}(X_i)), \\
P(X_i, X_j) = P(X_j \mid \text{Pa}(X_j)) \cdot P(X_i \mid \text{Pa}(X_i)).
\end{gather}
Then, assuming an intervention on $X_s$ and the truncated factorization formula (causal factorization with applied intervention \citep{pearl2009causality}), we write \Cref{eq:causal_factorization} for $X_i \to X_j$,
\begin{gather*}
\label{eq:truncated_factorization}
P(X \backslash X_s \mid \text{{do}}(X_s = x)) = \prod_{i \mid X_i \neq X_s} P(X_i \mid \text{{Pa}}(X_i)) \Big|_{x}, \\ 
P(X_i, X_j \mid \text{{do}}(X_s = x)) = \prod_{t \in \{i, \;j\}} P(X_t \mid \text{{Pa}}(X_t)) \Big|_{x},
\end{gather*}
where it becomes clear that altering $X_s$'s distribution directly affects its descendants. \hspace{1pt} $\square$

\begin{figure}
\begin{center}
\includegraphics[width=\linewidth]{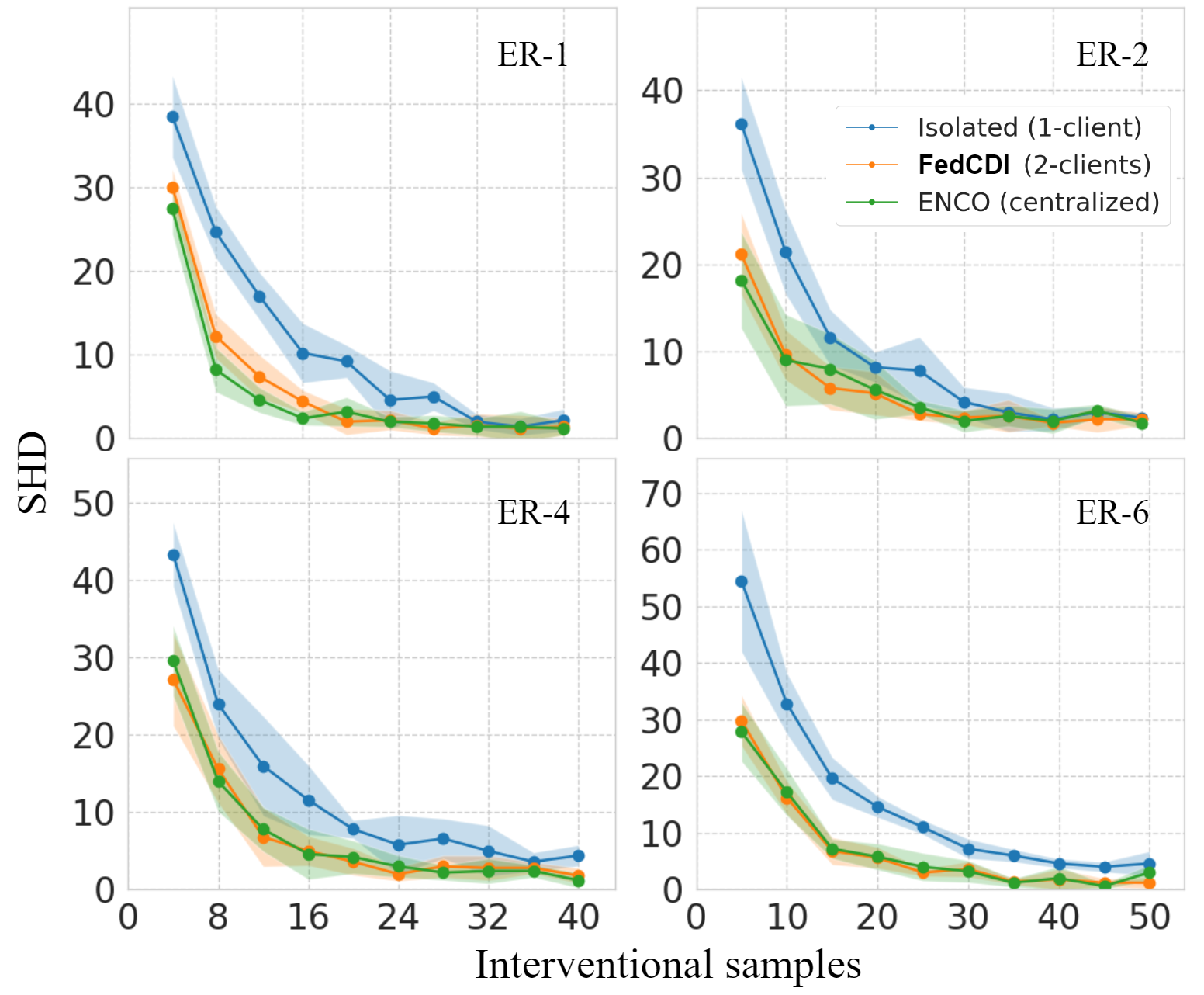}
\caption{\ours{} is rigorously compared against a centralized approach and an isolated client, taking for various interventional data sizes. Within the centralized approach, the entire dataset is available, while in both \ours{} and the isolated client, each client is limited to only half of $D$ distributed horizontally.}
\label{fig:balanced_int_aggregation_dataset}
\end{center}
\end{figure}

In order to find $r_{ij}^k$ for each client, we inject hypothetical mass $m_{s}^k$ to the node $X_s$ and let the mass flow in the DAG without its magnitude being divided by the fan-out of each node. When passing through any edge $X_q \to X_p$ on the way, the mass value is multiplied by $\psi_{qp}$ where $\psi_{qp}$ is the current aggregated belief in the network. The new mass then continues to flow in the graph in the same manner. For instance, the reliability score of the edge $X_s \to X_p$ is $\psi_{sp}^k m_{s}^k$. The score becomes $ \psi_{sp}^k \psi_{pq}^k m_{s}^k$ for $X_p \to X_q$. Should different mass values reach the same node from multiple paths, we pick the maximum regardless of the source variables. After a sufficient number of steps, the mass flows through the entire DAG.

\begin{figure}
\begin{center}
\includegraphics[width=0.99\linewidth]{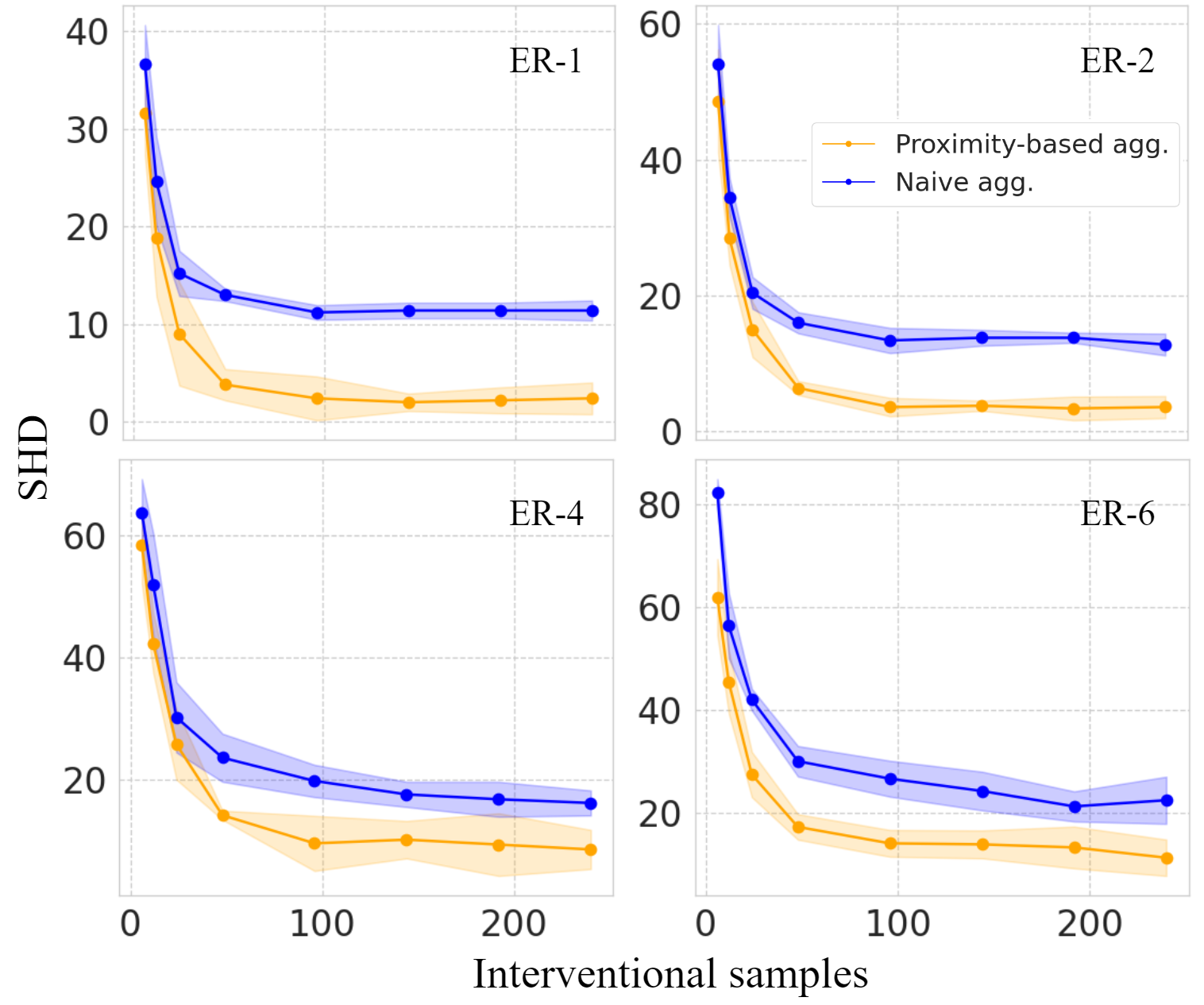}
\caption{We apply the proximity-based and naive aggregation methods to vertically distributed $D_{\mathcal{I}}$. Each curve corresponds to a 5-client setup, where clients have access to disjoint and covering subsets of vertically distributed interventional dataset, i.e., local datasets contain interventional samples on only 4 out of 20 dataset features.}
\label{fig:naive_vs_locality_aggregation}
\end{center}
\end{figure}
All clients perform the same process only once at each round, based on the latest aggregated belief, $\psi$. Let the mass passing through an edge be the reliability score for that edge. The process yields a set of reliability scores, $\{r^{1}_{ij}, r^2_{ij}, \ldots, r^K_{ij}\}$, assigned to each edge $X_i \to X_j$. We cannot directly use $r^k_{ij}$ values for aggregating the local beliefs $\{\psi_{ij}^1, \psi_{ij}^2, \ldots , \psi_{ij}^K\}$ as $\sum_{k} r_{ij}^k\neq 1$. To turn $r^k_{ij}$ values into probabilities, we apply a softmax function and write the new aggregated belief, $\forall k \in \{1, \ldots, K\}$,
\begin{equation}
\label{eq:reliability_calculation}
        \psi_{ij}^* = \sum_{k\in [K]}\hat{r}_{ij}^k \psi_{ij}^k, \quad
        \hat{r}_{ij}^k = \frac{e^{\beta r_{ij}^k}}{\sum_{k\in[K]}e^{\beta r_{ij}^k}},
\end{equation}
where $\psi_{ij}^*$ is the aggregated belief and $\beta > 0$ is the temperature parameter of the softmax function. The rate parameter $\beta$ controls the sensitivity of the distribution of $\hat{r}_{ij}^k$ to the difference of the reliability scores calculated by different clients. This procedure has the following parameters: 
\begin{itemize}
    \item $m_{i}^k, \; \forall i \in \{i \mid X_i \in X^{k}_{\mathcal{I}}\}$, the initial mass of $C_k$. We set this parameter proportionate to the number of interventional samples for $X_i$ available in $D^k$.
    \item $\beta$, the softmax temperature, optimized by a grid search over a computationally feasible set.
\end{itemize}

\Cref{alg:federated_setup_overview} presents \ours{} and our novel proximity-based aggregation. We conclude this section by providing a toy example to illustrate the proposed belief aggregation. 
\begin{example} Let's consider a 2-client setup. $C_1$ has interventional samples on $X_1$ (left), whereas $C_2$ holds interventional data on $X_3$ (right). We assume the clients have an equal number of samples, $m_1^1=m_2^3=1$, and $\beta = 2$ and $\psi_{ij}$ acquired from last aggregation are written on the edges. Edges with $\psi_{ij} < 0.5$ are omitted for simplicity. 

\begin{center}
  \scalebox{0.7}{
  \begin{tikzpicture}
  \node[circle, draw, fill=red!30] (X1) at (0,0) {X$_{1}$};
  \node[circle, draw, fill=blue!30] (X2) at (-1,-1) {X$_{2}$};
  \node[circle, draw, fill=blue!30] (X3) at (1,-1) {X$_{3}$};
  \node[circle, draw, fill=blue!30] (X4) at (3,-1) {X$_{4}$};
  \node[circle, draw, fill=blue!30, label=above:{\small $C_1: m_1^1=1$}] (X5) at (2,0) {X$_{5}$};
  \node[circle, draw, fill=blue!30] (X6) at (4,0) {X$_{6}$};

  \draw[->] (X1) -- (X2) node[pos=0.5] {0.6};
  \draw[->] (X1) -- (X3) node[pos=0.5] {0.6};
  \draw[->] (X3) -- (X4) node[pos=0.5] {0.5};
  \draw[->] (X3) -- (X5) node[pos=0.5] {0.9};
  \draw[->] (X5) -- (X6) node[pos=0.5] {0.8};
  \draw[->] (X4) -- (X6) node[pos=0.5] {0.7};
  \end{tikzpicture}
  \begin{tikzpicture}
  \node[circle, draw, fill=blue!30] (X1) at (0,0) {X$_{1}$};
  \node[circle, draw, fill=blue!30] (X2) at (-1,-1) {X$_{2}$};
  \node[circle, draw, fill=red!30] (X3) at (1,-1) {X$_{3}$};
  \node[circle, draw, fill=blue!30] (X4) at (3,-1) {X$_{4}$};
  \node[circle, draw, fill=blue!30, label=above:{\small $C_2: m_2^3=1$}] (X5) at (2,0) {X$_{5}$};
  \node[circle, draw, fill=blue!30] (X6) at (4,0) {X$_{6}$};

  \draw[->] (X1) -- (X2) node[pos=0.5] {0.5};
  \draw[->] (X1) -- (X3) node[pos=0.5] {0.5};
  \draw[->] (X3) -- (X4) node[pos=0.5] {0.8};
  \draw[->] (X3) -- (X5) node[pos=0.5] {0.9};
  \draw[->] (X5) -- (X6) node[pos=0.5] {0.5};
  \draw[->] (X4) -- (X6) node[pos=0.5] {0.5};
  \end{tikzpicture}}
\end{center}
The flow of mass for $C_1$ and $C_2$ yields:
\begin{gather*}
    r_{5,6}^1 = 0.6 \cdot 0.9 \cdot 0.8 \cdot m_1^1 = 0.432, \\    
    r_{5,6}^2 = 0.9 \cdot 0.5 \cdot m_3^2 = 0.540 \Rightarrow R_{5,6} = \{0.432, 0.540\}.
\end{gather*}
Then, the final reliability score of $C_1$ and $C_2$ for $X_5 \to X_6$ is calculated by: 
\begin{gather*}
    \hat{r}_{5,6}^1 = \frac{e^{(2 \times 0.432)}}{e^{(2 \times 0.432)} + e^{(2 \times 0.540)}} \approx 0.446, \\ \hat{r}_{5,6}^2 = \frac{e^{(2 \times 0.540)}}{e^{(2 \times 0.540)} + e^{(2 \times 0.432)}} \approx 0.554.
\end{gather*}
\end{example}

\section{Experiments}
\label{sec:experiments}

\begin{figure}
\begin{center}
\includegraphics[width=0.98\linewidth]{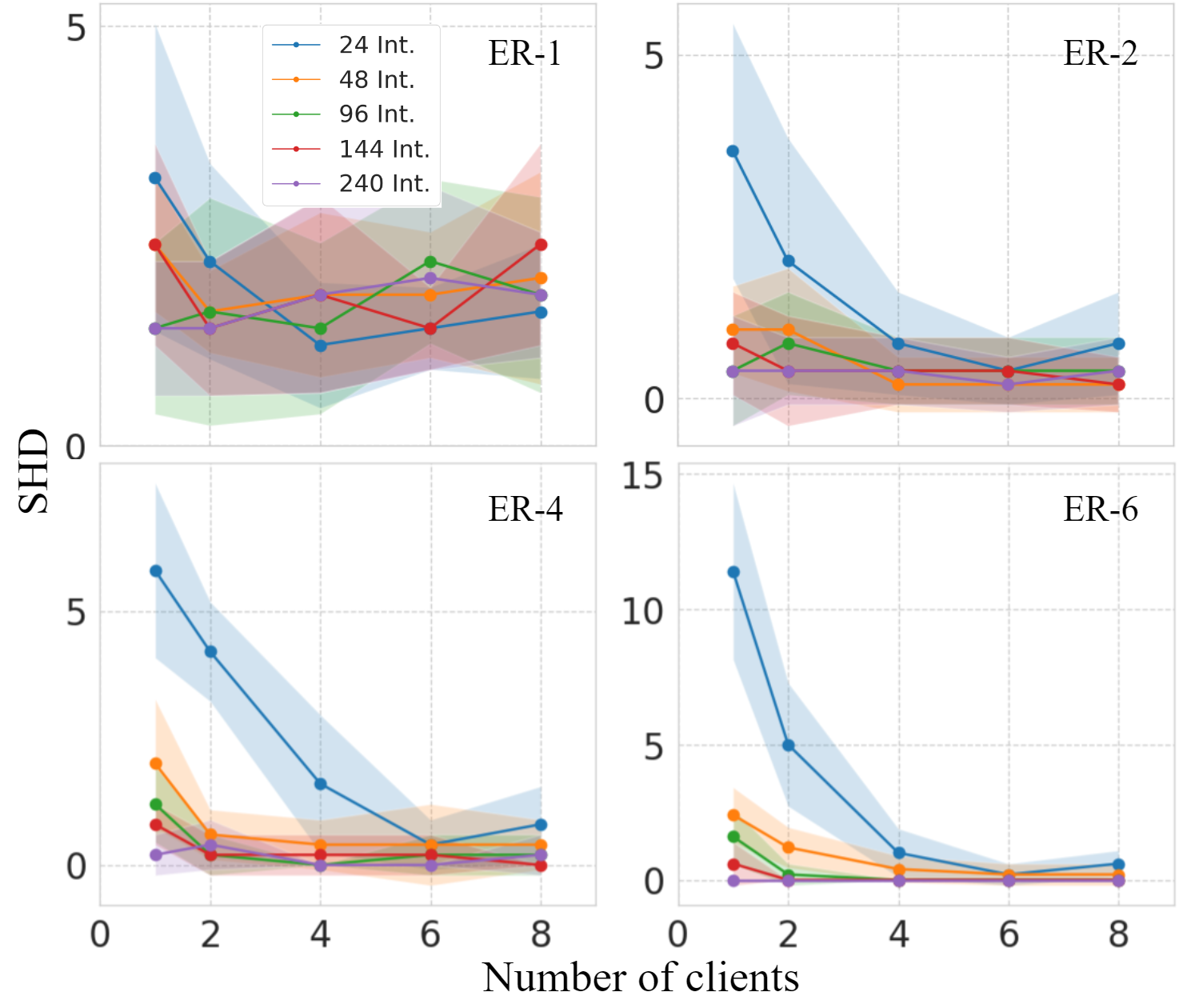}
\caption{Effect of additional clients participating in \ours{} for different $D_{\mathcal{I}}$ sizes. The discovered DAG is more accurate where $size(D_{\mathcal{I}})$ increases with the addition of new clients. }
\label{fig:client_sweep_nodiv}
\end{center}
\end{figure}

To empirically showcase the efficacy of \ours{}, we conduct a thorough evaluation by comparing its results against both a centralized setting and non-collaborating clients. Additionally, we scrutinize naive averaging-based and proximity-based aggregation methods discussed earlier, providing specific insights into the limitations of the former when confronted with vertical data distributions. Finally, we conclude our assessment with scalability experiments and a comparative analysis of \ours{} against selected baselines in the literature on both real-world and synthetic graphs. 

\begin{figure}
\begin{center}
\includegraphics[width=\linewidth]{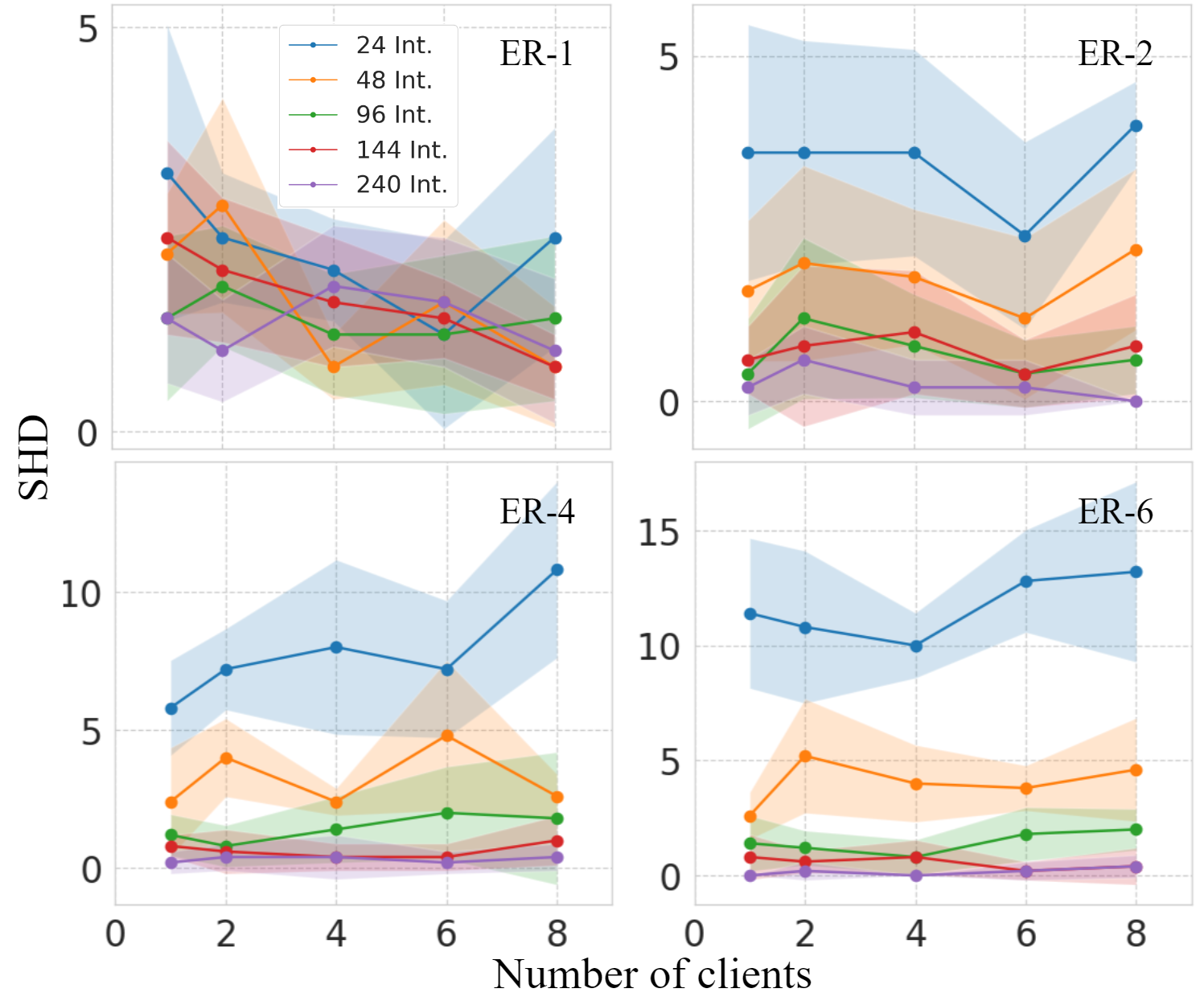}
\caption{\ours{} manages to sustain the knowledge when the fixed-size $D_{\mathcal{I}}$ is further divided by the increase in number of clients, as opposed to \Cref{fig:client_sweep_nodiv}.}
\label{fig:client_sweep_div}
\end{center}
\end{figure}

\subsection{Experiment setup}
\label{sec:experiments_setup}

\textbf{DAGs.} For \emph{synthetic} data, we employ the random Erdős–Rényi (ER) model to pick experimental DAGs of size $d = 20$, where each edge is added with an independent pre-determined probability from the others~\citep{erdos2011evolution}. We experiment on data generated by $ER-n$ DAGs where $n \in \{1, 2, 4, 6\}$. According to the definition, the value of $n.d$ is equal to the expected number of edges. For the \emph{real-world} graphs in the last section, we turn to \texttt{bnlearn} package \citep{bnlearn}, and pick Sachs \citep{sachs2005causal}, Alarm \citep{beinlich1989alarm}, and Asia \citep{lauritzen1988local} graphs.

\begin{figure}
\includegraphics[width=\linewidth]{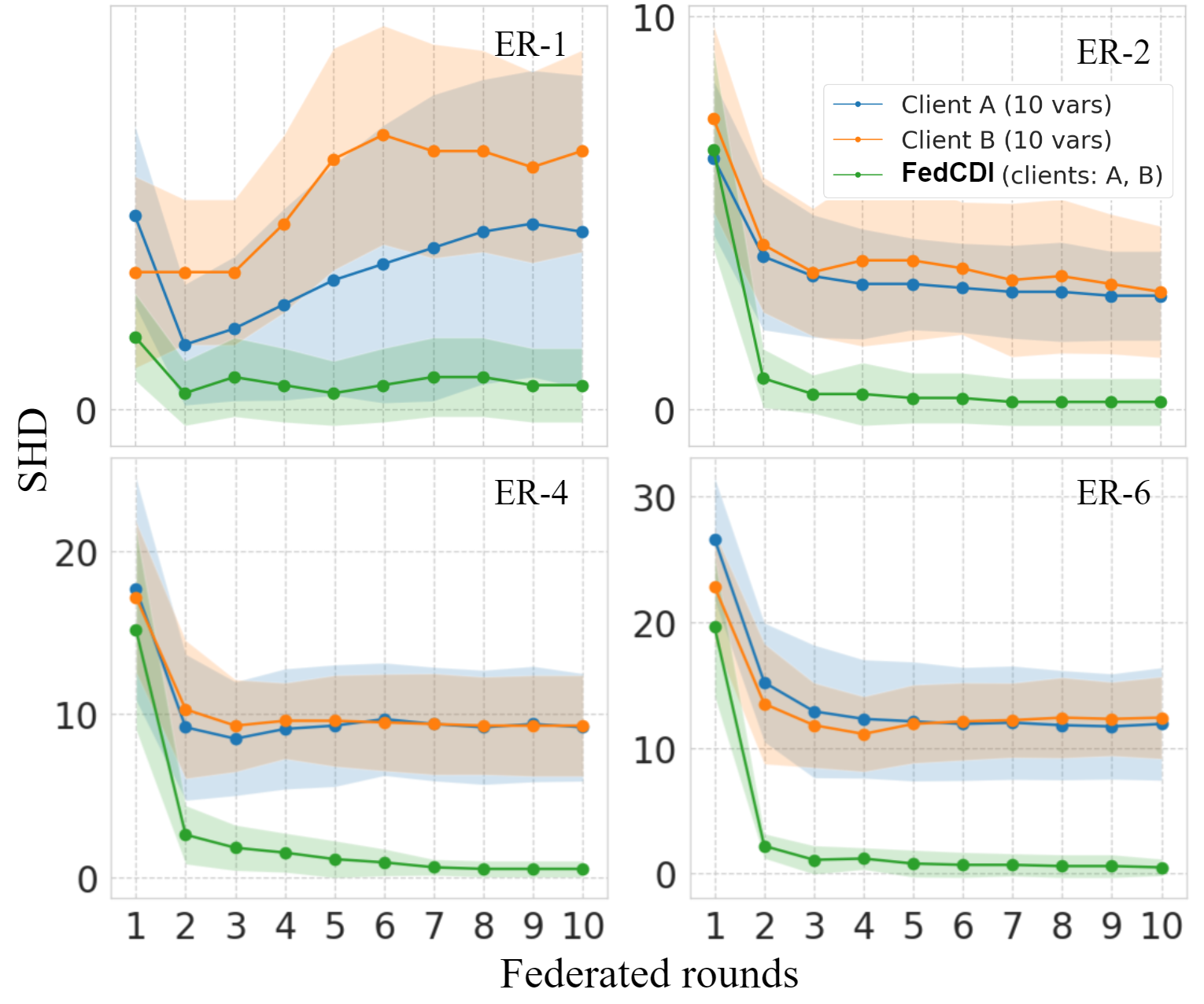}
\caption{Non-collaborative clients A and B with access to vertical subsets of $D_{\mathcal{I}}$, and \ours{} where the same A and B clients now share their belief about the structure.}
\label{fig:unbalanced_int_10_10}
\end{figure}

\textbf{Dataset.} The global dataset, $D$, is generated based on the previously sampled DAGs, and by employing the a synthetic data generation mechanism similar to \citet{ke2020dependency}. The process utilizes randomly initialized neural networks with two layers with categorical inputs to model the ground-truth conditional distributions. We sample categorical data with 10 categories in different sizes for both interventional and observational sets. For all experiments, the observational data is horizontally distributed among clients. However, a client may receive interventional samples for all intervened variables (horizontal) or only a subset of those variables (vertical). Further details about the data generation process appear on \Cref{appendixb:detailed_experimental_setup}. We also apply \ours{} to the data generated from structured graphs in \Cref{appendixa:supporting_experiments}.

\textbf{Evaluation.} We compare the outcome of each setup to the ground truth by calculating the Structural Hamming Distance (SHD) between the two. Note that the outcome for \ours{} is calculated by applying a binary step function (threshold = 0.5) to the final belief matrix. We then run each experiment for 20 random seeds, and report the SHD average and confidence interval.

\subsection{Horizontal data distribution} 
\label{sec:experiments_results_balanced}
We start with a scenario where both $D_{\mathcal{O}}$ and $D_{\mathcal{I}}$ are split horizontally. \Cref{fig:balanced_int_aggregation_dataset} presents multiple experiments with our method while employing the proximity-based aggregation. The results clearly demonstrate that two collaborative clients outperform an isolated client. Moreover, it is evident that \ours{} discovers a DAG with the same precision as the centralized approach. Therefore, the belief aggregation step performs an effective aggregation of clients' information without directly exchanging samples. Note that as the size of $D_{\mathcal{I}}$ increases in \Cref{fig:balanced_int_aggregation_dataset}, the results of the isolated client come closer to \ours{}, as each client now holds adequate samples to find the underlying DAG without collaboration.
\begin{figure}
\includegraphics[width=0.99\linewidth]{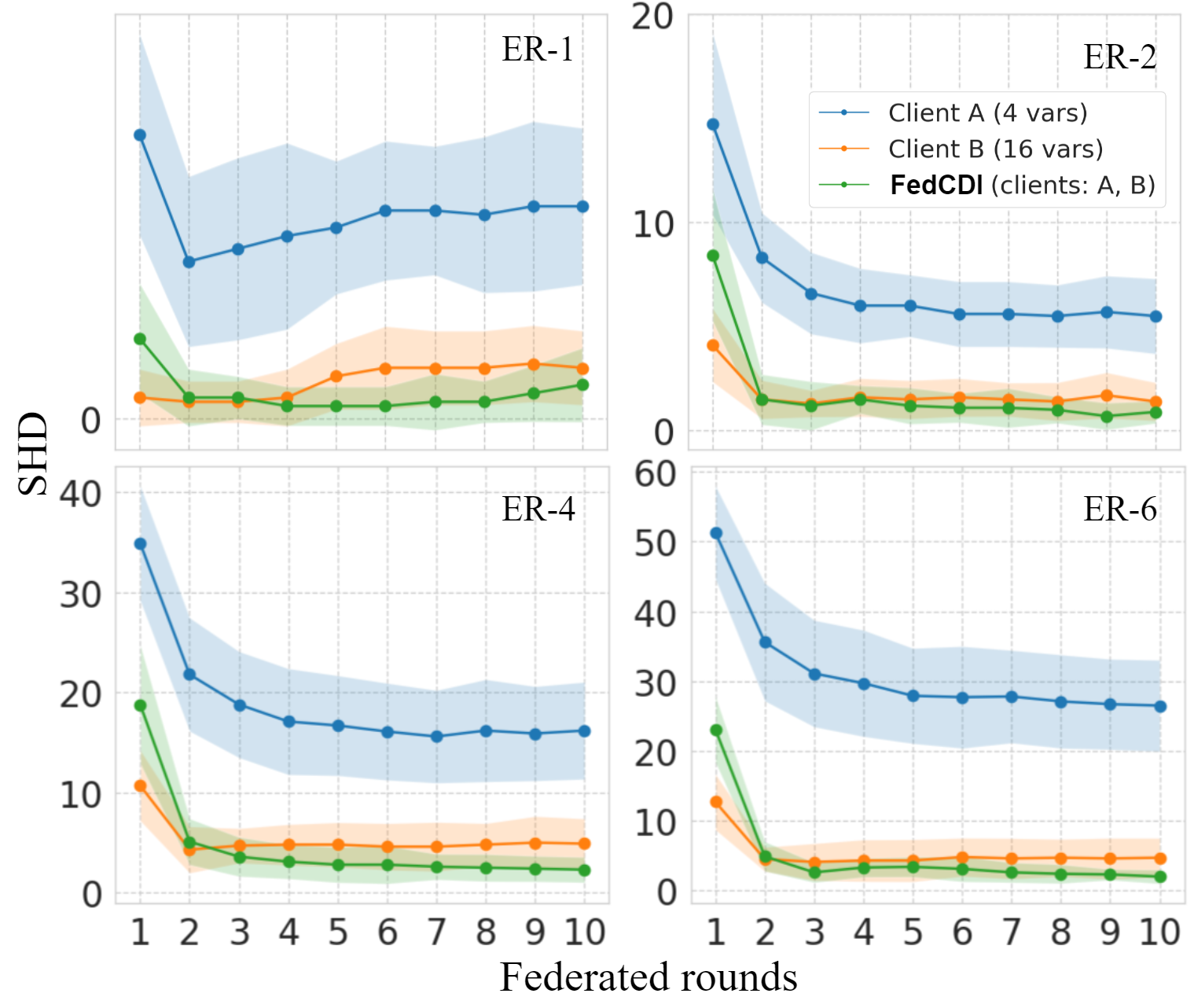}
\caption{We further extend the setup in \Cref{fig:unbalanced_int_10_10} to demonstrate the way our aggregation method leverages the information received from clients with fewer intervened covariates.}
\label{fig:unbalanced_int_16_4}
\end{figure}
\subsection{Vertical distribution of interventional data}
\label{sec:experiments_results_unbalanced}
To highlight the critical role of interventional data in FCD, we vertically distribute $D_{\mathcal{I}}$, limiting clients' interventional samples to a subset of dataset covariates. \Cref{fig:naive_vs_locality_aggregation} compares naive averaging-based aggregation method to the more sophisticated proximity-based aggregation and suggests that naive averaging underperforms as it disregards clients' unique access to interventions. 

We further showcase the ability of proximity-based aggregation in \Cref{fig:unbalanced_int_10_10}. Two isolated clients are compared against a 2-client federated setup, where the same two clients collaborate through \ours{} and yield superior results.  Moreover, \Cref{fig:unbalanced_int_16_4} illustrates that a client with access to the majority of interventional data still benefits from collaborating with the less significant client. This is justified as \ours{} exploits the second client's knowledge of interventional samples, which are critical to precise DAG discovery. 
\begin{table*}
\caption{Comparison between \ours{} and baselines selected in \Cref{sec:baselines_generalization_exps} for \textbf{synthetic} ($ER-n$) and \textbf{real-world} (\textit{Sachs}, \textit{Alarm}, \textit{Asia}) graphs. Results are averaged over 20 random seeds. The top two highest scores in each graph are bolded.} 
\label{tab:int_data_baselines}
\centering
\scalebox{0.89}{
\begin{tabular}{@{}c|ccccc|ccc@{}}
\toprule
\multicolumn{1}{c}{Data} &
\multicolumn{1}{c}{Method} & \multicolumn{1}{c}{ER-1} & \multicolumn{1}{c}{ER-2} & \multicolumn{1}{c}{ER-4} & \multicolumn{1}{c}{ER-6} & \multicolumn{1}{c}{Sachs} & \multicolumn{1}{c}{Alarm} & \multicolumn{1}{c}{Asia} \\ \midrule
 & GIES                           & $14.4 \pm 2.1$                     & $46.9 \pm 0.8$                     & $127.3 \pm 1.6$                         & $168.8 \pm 0.7$                         
& $21.3 \pm 4.3$ & $43.2\pm 7.5$ & $13.8 \pm 3.1$   \\

\multirow{1}{*}{Centralized \&} & IGSP                            & $11.8 \pm 2.8$                     & $43.7 \pm 0.7$                         & $118.5 \pm 0.4$                         & $152.8 \pm 0.1 $                        & $14.4 \pm 3.4$ & $25.1 \pm 1.4$ & $12.6 \pm 2.7$            \\

\multirow{1}{*}{Interventional} & DCDI                           & $3.0 \pm 1.1$                       & $24.3 \pm 2.1 $                         & $31.4 \pm 8.6$                          & $43.0 \pm 12.3$                           & $6.9 \pm 1.2$ & $36.8 \pm 2.3$ & $6.1 \pm 1.6$      \\

 & ENCO                            & $\mathbf{1.9 \pm 0.1}$                       & $4.3 \pm 0.1 $                         & $5.7 \pm 0.2$                          & $7.8 \pm 0.3$  
& $\mathbf{0.5 \pm 0.4}$ & $13.8 \pm 2.5$ & $2.0 \pm 0.1$  
     
 \\ \midrule 

\multirow{1}{*}{Federated \&} & FedDAG                              & $10.3 \pm 1.9$                       & $14.2 \pm 1.4$                         &$17.6 \pm 0.7 $                          & $24.5 \pm 3.6$            & $5.4 \pm 1.0$ & $11.2 \pm 0.8$ & $\mathbf{1.5 \pm 0.6}$                          

         \\

\multirow{1}{*}{Observational} & NOTEARS-ADMM                             & $7.7 \pm 3.0$                       & $15.8 \pm 4.9$                         &$17.2 \pm 2.4 $                          & $21.1 \pm 5.3$            & $6.1 \pm 2.2$ & $\mathbf{8.2 \pm 1.5}$ & $4.7 \pm 1.2$   
\\ \midrule

\multirow{1}{*}{Federated \&} & \textbf{\ours{} (2-clients)}                        & $2.6 \pm 0.6$                       & $\mathbf{3.9 \pm 0.4}$                         &$\mathbf{4.1 \pm 0.1 }$                          & $\mathbf{5.5 \pm 0.3}$            & $0.9 \pm 0.3$ & $19.8 \pm 2.6$ & $8.8 \pm 2.5$                            \\ 

\multirow{1}{*}{Interventional} & \textbf{\ours{} (4-clients)}                        & $\mathbf{2.3 \pm 0.4}$                       & $\mathbf{3.3 \pm 0.7}$                         &$\mathbf{4.6 \pm 0.6 }$                          & $\mathbf{5.7 \pm 0.4}$            & $\mathbf{0.8 \pm 0.5}$ & $\mathbf{8.0 \pm 3.1}$ & $\mathbf{1.7 \pm 0.4}$                            \\
\bottomrule
\end{tabular}
}
\end{table*}
\subsection{Performance with more collaborating clients}
\label{sec:more_clients_performance_exps}
To assess the effect of increasing the number of clients, we performed two series of experiments. In \Cref{fig:client_sweep_nodiv}, new clients join \ours{} by bringing their own local dataset. Therefore, the size of $D_{\mathcal{I}}$ increases upon adding a new client, and more information becomes available in the network, while $D^k$ have a fixed size. Then, we keep $size(D_{\mathcal{I}})$ constant in \Cref{fig:client_sweep_div}, and then increase the number of clients. Each new client receives a share by joining the setup, leaving the others with fewer samples.

The results implicate that \ours{} experiences higher precision when clients with new information are added to the collaborative environment. Each client brings fresh samples, which in turn helps the aggregation stage to learn more accurate existence probabilities. However, even in the case of $D_{\mathcal{I}}$'s further division by adding new clients, \ours{} maintains effective knowledge aggregation. The only breaking point happens when clients are left with so few samples, that even the LCDM itself fails to converge to a local graph (see convergence conditions in \citet{lippe2021efficient}).

\subsection{Baselines and generalization} 
\label{sec:baselines_generalization_exps}
We compare our setup against well-known DAG discovery methods which leverage both interventional and observational data. From the methods listed in \citet{guo2020survey}, we pick GIES \citep{hauser2012characterization}, IGSP \citep{wang2017permutation}, and DCDI \citep{brouillard2020differentiable}, apply them in a centralized manner to $\{D_{\mathcal{O}}, D_{\mathcal{I}}\}$, and report the results. We then apply  FedDAG~\citep{gao2021feddag} and NOTEARS-ADMM~\citep{ng2022fed-bayesian-nets} to the same data, despite their sole reliance on observational samples. The dataset size is the same across all baselines, but samples are divided between clients in \ours{}, FedDAG, and NOTEARS-ADMM. The results indicate the capability of \ours{} in knowledge aggregation, considering that each client has access to less information compared to centralized approaches, yet still the setup as a whole manages to yield precise discoveries.


\section{Limitations and Future Work}
\label{sec:limitations_future_work}

\textbf{Real-world data.} Real-world causal discovery datasets lack the required interventional samples to enable \ours{}'s capabilities. fMRI Hippocampus~\citep{poldrack2015long} is an example of datasets employed by federated DAG discovery methods which solely contain observational samples. Even though we use real-world graphs from \texttt{bnlearn}~\cite{bnlearn}, we are bound to generate the corresponding data same as previous work~\citep{ke2020dependency, brouillard2020differentiable, lippe2021efficient} in the literature. 

\textbf{Client data limitation.} We observed that if clients have a limited local dataset lower than a certain threshold, the resulting DAG becomes highly inaccurate. Though this phenomenon is intuitive, the threshold depends on the choice of LCDM and the underlying data generation processes. Future work could propose a generic method to yield this threshold. 

\textbf{Computational complexity.} Although running clients' LCDM in parallel potentially reduces the overall computational time compared to a centralized approach, it can pose a higher computational load depending on the nature of the learning method. For instance, if the computational cost of the LCDM is linear regarding data and the overheads are negligible, then the computational load remains the same as a centralized approach. On the other hand, having a large overhead aside from dataset size leads to higher computational load compared to centralized approaches.  

\textbf{Acyclicity.} One must ensure that the adjacency matrix obtained after several rounds of aggregation is still a DAG. Pruning the adjacency matrix corresponding to the final estimation is one option. Yet, future work could integrate extra constraints in the aggregation process to ensure acyclicity.

\section{Conclusion}
\label{sec:conclusion}
We developed a novel approach to federated DAG discovery from distributed datasets, aiming to enhance client privacy by avoiding the sharing of local samples. Concentrating on interventional data, we successfully uncover the underlying causal structure with precision and effectively handle both homogeneous and heterogeneous distributions of interventional data. Through extensive experiments, we demonstrate the scalability and performance of \ours{}, and compare it to centralized and federated methods. Our use of a naive aggregation further highlights the advantage of considering clients' interventions. \ours{} empirical results underscore its superiority against compared to previous FCD methods.

\section{Reproducibility}
\label{sec:reproducibility}
\ours{}'s source code (\href{https://sites.google.com/view/fedcdi/home}{sites.google.com/view/fedcdi}) is published, and the repository includes all the essential reproducibility instructions. To ease off reproducibility, the raw data can be downloaded, especially for experiments which require significant computational resources.


\newpage
\bibliography{references}

\onecolumn
\title{Federated Causal Discovery from Interventions\\(Supplementary Material)}
\maketitle

\pagenumbering{arabic}

\appendix
\section{Supporting Experiments}
\label{appendixa:supporting_experiments}
We offer a thorough series of additional experiments utilizing the \ours{} framework, designed to rigorously assess its efficacy from diverse perspectives. These experiments encompass evaluations with both the random ER graphs outlined in the paper and structured DAGs commonly encountered in the literature.

Please note that throughout all experiments, the experimental criteria and metrics remain consistent with those outlined in the main paper, unless explicitly stated otherwise. We also use our proximity-based intervention aggregation method for all experiments, and the data generation mechanisms are the same for both ER and structured graphs.


\subsection{Performance of federated rounds}
\begin{figure}[htbp]
\centering
\includegraphics[width=0.8\linewidth]{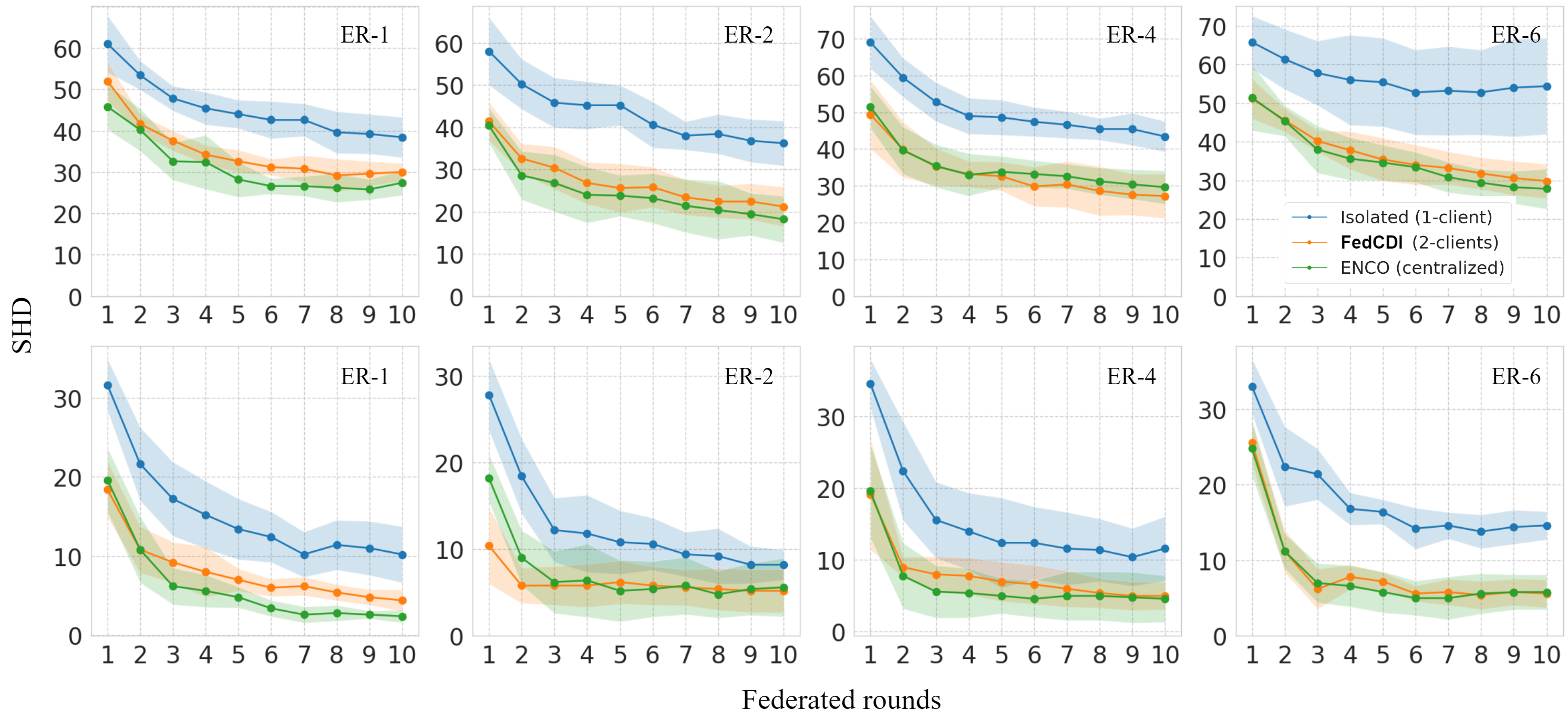}
\caption{We provide a closer look into the gradual improvement at each federated round. Our method is compared against a centralized approach and an isolated client for two $D_{\mathcal{I}}$ sizes: 4 (first row) and 20 (second row). The centralized approach has access to the entire dataset, while each client in \ours{} and the isolated client all have access to half of the dataset. The plots demonstrate that \ours{} is closely following a centralized approach at each federated round.}
\label{fig:balanced_int_aggregation_rounds}
\end{figure}

\Cref{fig:balanced_int_aggregation_rounds} experiments demonstrate the way \ours{} performs after each \emph{federated round}. These experiments are conducted for two different dataset sizes, and are a special case of \Cref{fig:balanced_int_aggregation_dataset} in the main paper. The plots indicate the framework's improvement at the end of each round (aggregation step), and the resulting final convergence after a few rounds.  

\begin{figure}[!ht]
\includegraphics[width=\linewidth]{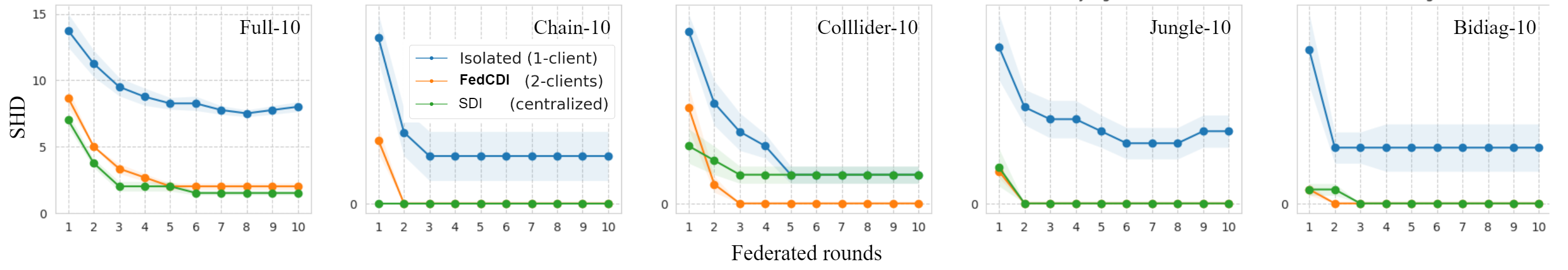}
\caption{Evaluation of our approach when using SDI as the LCDM, instead of ENCO. We use structured graphs of size 10 here, and the improvement caused by the federated setup is still observable for all structured graph types. We need to reduce the graph size due to SDI's computational time to achieve tractable experiments.}
\label{fig:sdi_federated_str_all}
\end{figure}

\subsection{Generalization to other LCDM candidates}
We believe that other continuous-optimization based structure discovery methods can be employed as the LCDM in our approach if modifications explained in our Methodology are applicable. For instance, in \Cref{fig:sdi_federated_str_all}, we switch the LCDM from ENCO to SDI~\citep{ke2020dependency}, a work on causal discovery from observational and interventional samples. The SDI method is an iterative, score-based, continuous optimization strategy with three interconnected phases. Throughout these stages, both a functional representation of a set of independent causal processes and a structural representation of a DAG are trained simultaneously until convergence. Since the structural and functional parameters are interdependent, they undergo alternating stages of training. This alternation occurs between fitting the graph to observational data and scoring the graph on interventional samples, ultimately resulting in a belief about the underlying DAG.

Even when employing a federated setup, the results still demonstrate improvement upon replacing ENCO with SDI. It's important to note that due to the significant computational demands of SDI, our experimentation is limited to structured graphs with 10 nodes. This computational challenge is why we primarily utilize ENCO in the main text and for most experiments. However, we anticipate that future advancements in implementation and integration of SDI into the federated setup could mitigate these complexities. Additionally, it's worth mentioning that the underlying graphs for these experiments differ from the ER graphs introduced in the main text. Further details about the underlying data generator graphs for these experiments can be found in \Cref{appendixc:additional_experiments}.

\subsection{Analysis of the prediction entropy}
\label{appendixa:entropy_exps}

\emph{Consensus}, in the context of distributed systems, denotes the collective agreement among processes to converge on a specific value, typically attained through the voting mechanism of each process. This concept holds significant importance in federated learning. In this decentralized environment, achieving consensus ensures that all participating clients synchronize their contributions to the model update, resulting in a consistent and accurate global model. Consensus mechanisms play a crucial role in addressing the challenge of model divergence, where variations in local data and training conditions among clients can lead to divergent models. By attaining consensus, federated learning effectively mitigates the effects of data heterogeneity, ensuring that the aggregated model reflects a shared understanding of the data.

\begin{figure}[!ht]
\centering
\includegraphics[width=0.7\linewidth]{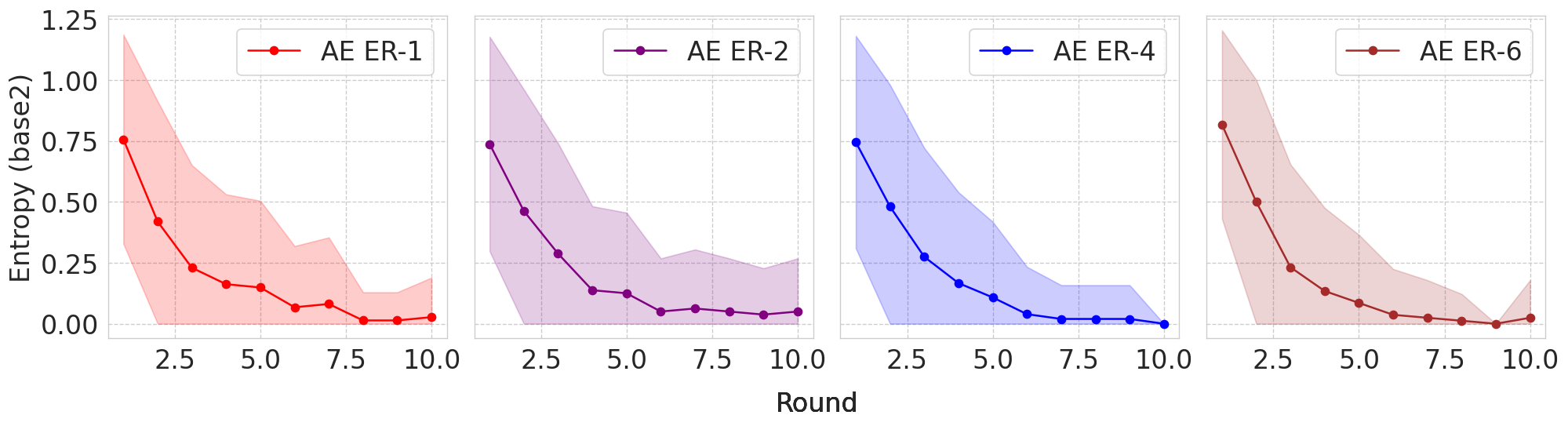}
\caption{Average edge entropy for all the elements in the clients' adjacency matrix for a 10-client federated setup is plotted here. The results reveal a substantial decrease in the values of edge entropy after each round, which serves as an empirical sign of convergence and indicates that the federated setup is progressing towards reaching a consensus among clients over the federated rounds. This result is critical to eliminate clients' uncertainty about the global DAG.}
\label{fig:entropy_graphs}
\end{figure}

\begin{figure}[!ht]
\centering
\includegraphics[width=\linewidth]{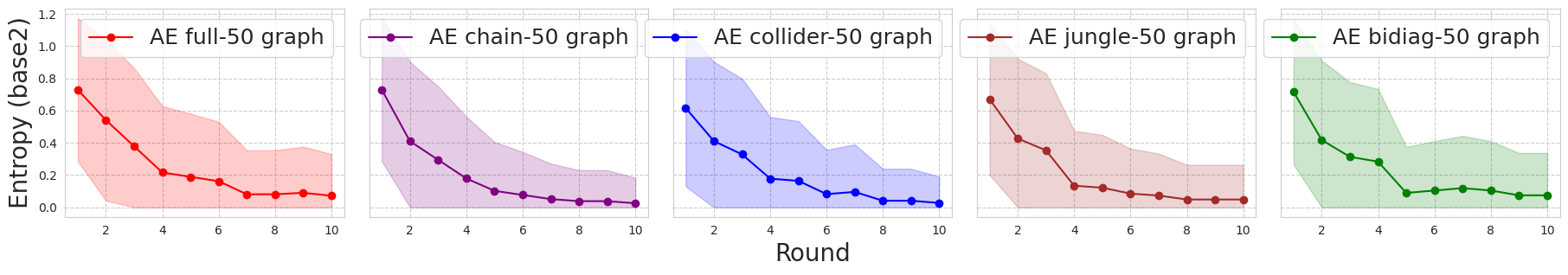}
\caption{Average edge entropy for all the elements in the clients' adjacency matrix for a 10-client federated setup plotted for structured graphs. The results reveal the same levels of decrease in the values of edge entropy after each round as \Cref{fig:entropy_graphs}. The decreasing trend resembles the way clients with no access to close intervened covariates yield to the ones with more valuable knowledge over time.}
\label{fig:entropy_graphs_str}
\end{figure}

In our study, we assess clients' consensus through calculating the \emph{average entropy} (AE) for each edge. This measure is based on the information exchanged between clients and the server at the end of each round. The empirical results depicted in \Cref{fig:entropy_graphs} and \Cref{fig:entropy_graphs_str} showcase the dynamics of average entropy reduction across different causal graphs as the rounds progress. Remarkably, our proposed approach demonstrates the ability to attain consensus in most cases within 10 rounds, although in certain scenarios, we observe a substantial reduction in average entropy without reaching complete consensus. These findings highlight the effectiveness of our approach in reducing uncertainty and aligning the distributed processes towards a shared belief about the global causal structure.

\subsection{Experiments on structured graphs}
\label{appendixc:additional_experiments}
We conduct further experiments on a range of structured graphs to further evaluate \ours{}. These graphs include the \textit{chain}, \textit{bidiag}, \textit{full}, \textit{collider}, and \textit{jungle} graphs. In the \textit{bidiag} graphs, the probability of a node $X_i$ being connected to its immediate parents $X_{i-1}$ and $X_{i-2}$ is defined as $P(X_i=1|X_{i-1}=1, X_{i-2}=1)$. Similarly, the \textit{chain} graph represents a simplified version, where the probability of a node $X_i$ being connected to its immediate parent $X_{i-1}$ is defined as $P(X_i=1|X_{i-1}=1)$. Collider graphs involve a specific node $X$ having all other nodes as its parents, with the probability of a node $X_i$ being connected to its parents being $P(X_i=1|Pa(X_i))$. The \textit{full} graph represents the densest connected graph possible, where the probability of a node $X_i$ being connected to all preceding nodes $X_1, X_2, \ldots, X_{i-1}$ is defined as $P(X_i=1|X_1=1, X_2=1, \ldots, X_{i-1}=1)$. Lastly, the jungle graph introduces a binary tree structure with additional connections to a node's parent's parent. These experiments provide valuable insights into how our approach performs under different graph configurations, capturing diverse causal relationships and their complexities. 

\begin{figure}[!ht]
\centering
\includegraphics[width=\linewidth]{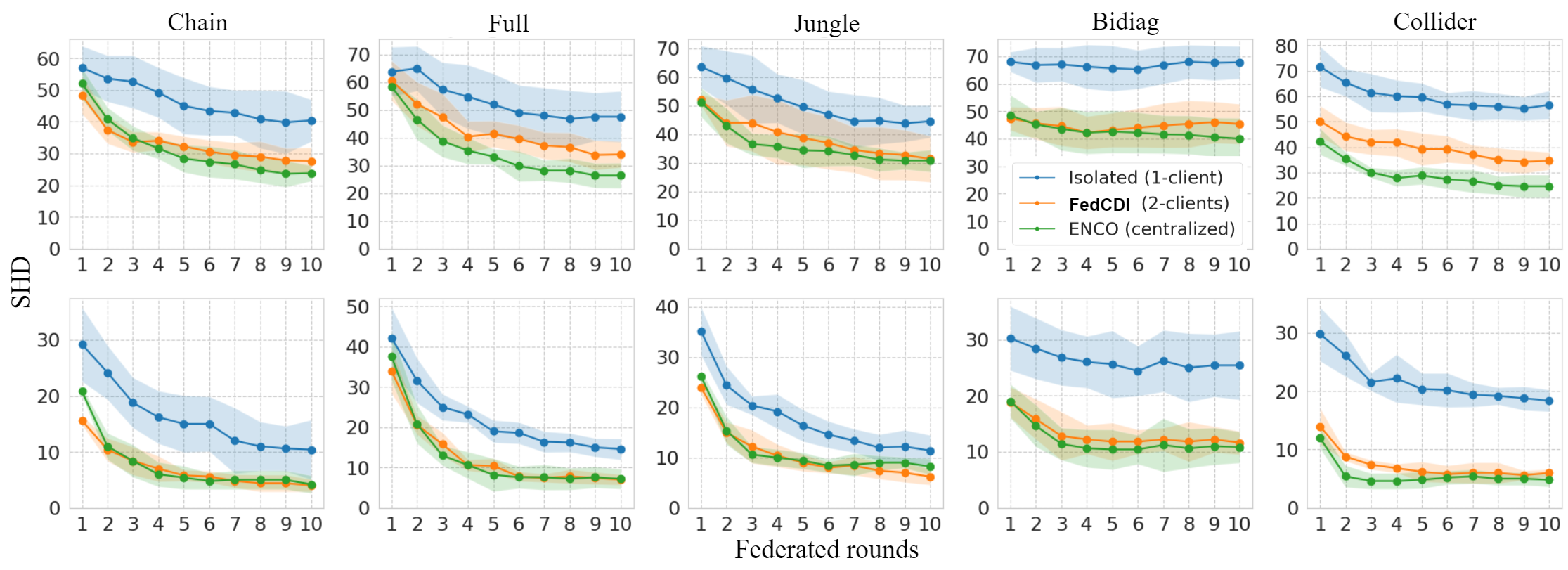}
\caption{We provide a detailed analysis of the incremental improvement achieved in each federated round for \emph{structured graphs}. Our method is compared against a centralized approach and an isolated client setting using two different sizes of $D_{\mathcal{I}}$: 4 (first row) and 20 (second row). In the centralized approach, the entire dataset is available, while in both \ours{} and the isolated client, each client has access to only half of the dataset. The plots clearly illustrate that \ours{} consistently tracks the performance of the centralized approach in every federated round.}
\label{fig:balanced_int_aggregation_rounds_str}
\end{figure}

\begin{figure}[!ht]
\centering
\includegraphics[width=\linewidth]{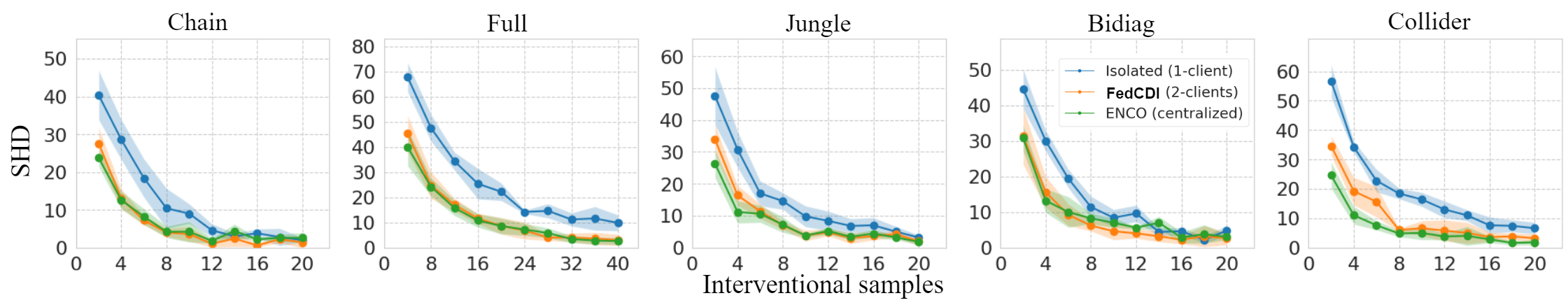}
\caption{Our method is compared against a centralized approach and an isolated client based on the size of interventional data. The centralized approach has access to the entire dataset, while each client in \ours{} and the isolated client all have access to half of the dataset. The results of comparison to the centralized approach with structured graphs are aligned with that of the random ER graphs.}
\label{fig:balanced_int_aggregation_dataset_str}
\end{figure}

\begin{figure}[!ht]
\centering
\includegraphics[width=\linewidth]{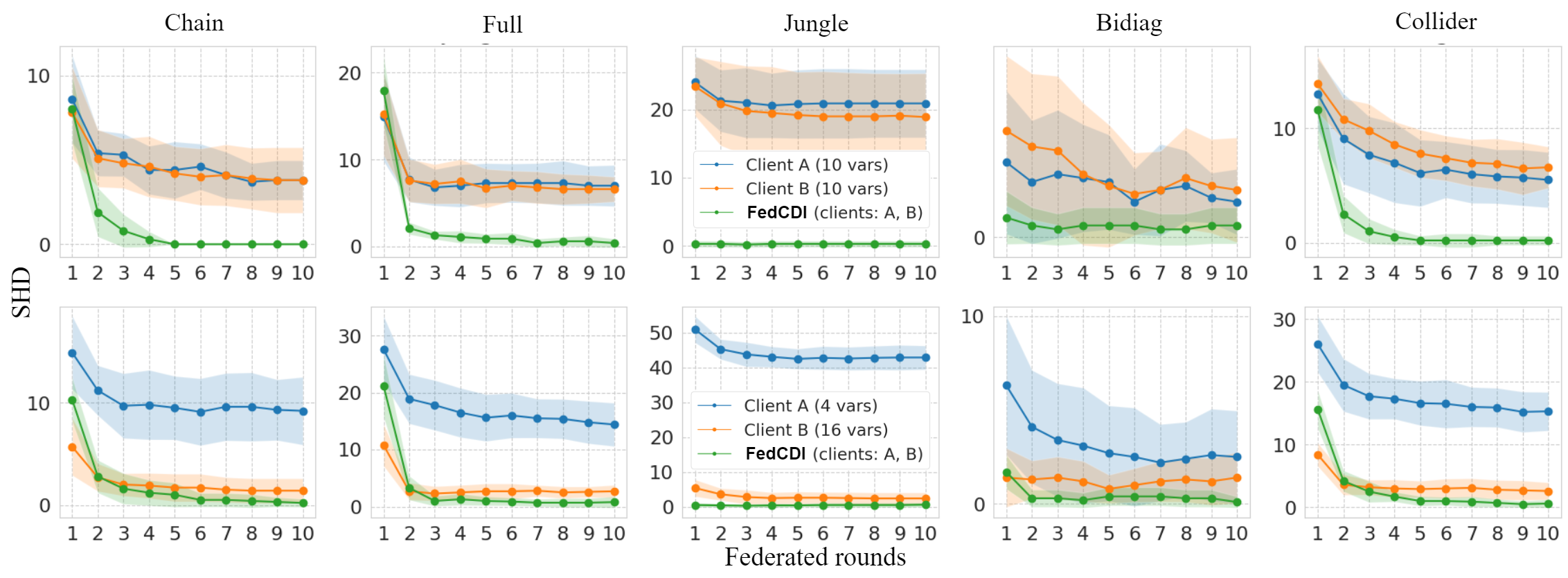}
\caption{We compare three different setups with vertical distribution of $D_{\mathcal{I}}$, this time considering structured graphs. The setups are non-collaborative clients A and B with access to only a vertical subset of $D_{\mathcal{I}}$, and a federated setup where both clients collaborate through \ours{}. Evidently, \ours{} can take advantage of clients' insight about the global structure, even from the less significant clients.}
\label{fig:unbalanced_int_aggregation_round_str}
\end{figure}

\Cref{fig:balanced_int_aggregation_rounds_str} and \Cref{fig:balanced_int_aggregation_dataset_str} show \ours{} performing consistently on datasets generated by structured graphs. Note that all the graphs here have the same size 20 as the random graphs, and the datasets are produced with the same process of \citet{lippe2021efficient} and \citet{ke2020dependency} with the same sizes as the main paper. The next plot, \Cref{fig:unbalanced_int_aggregation_round_str}, depicts the experimental results for vertical distribution of $D_{\mathcal{I}}$ where the underlying data generation process is built upon structured graphs, further verifying the results in the main paper.

\begin{figure}[!ht]
\includegraphics[width=\linewidth]{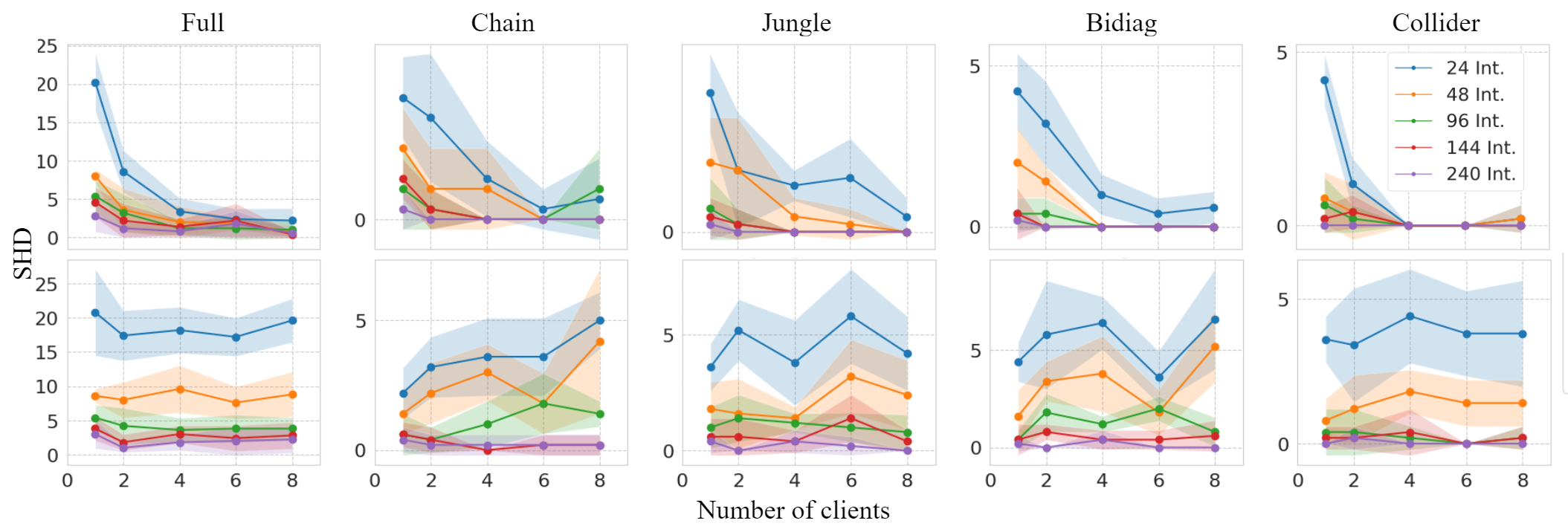}
\caption{In the initial row, we witness the impact of involving more clients in \ours{} across varying sizes of $D_{\mathcal{I}}$. As the number of clients grows, the resulting DAG becomes more precise, correlating with the enlargement of $size(D_{\mathcal{I}})$. Unlike the first row, our approach effectively maintains knowledge as the fixed-size $D_{\mathcal{I}}$ is divided further among clients, where no new information added to the network with the advent of new clients.}
\label{fig:add_clients_all_str}
\end{figure}

\section{Experiment Setup and Details}
\label{appendixb:detailed_experimental_setup}
This section is dedicated to further details related to our experiments. These include the size and distribution of synthetic data, and a detailed algorithm for implementation of our proximity-based aggregation method.
\subsection{Dataset size and distribution}
For the entire experiments, aside from the baselining table, we use $10000$ observational samples as $D_{\mathcal{O}}$ and $2000$ samples for $D_{\mathcal{I}}$. For the baselining, however, we increase the size of $D_{\mathcal{O}}$, $D_{\mathcal{I}}$ to $25k$ and $5k$ to make the experiments compatible with other methods such as IGSP. These numbers correspond to the size of $D$, that is, the global dataset yet to be distributed among several clients. For example, in a 2-client setup, each client typically gets half of the observational data. Depending on the vertical or horizontal split of the interventional data, a client might get access to samples from all or part of dataset features. Samples of interventional data corresponding to a specific feature are further divided between clients if they share an intervened covariate. In scenarios where clients have access to disjoint covariates, the entire interventional samples corresponding to the intervened covariate is accessible to the client with access to interventions on that feature. 

\subsection{Algorithm for aggregation method}
We provide a pseudocode for the proximity-based aggregation method in \Cref{alg:federated_setup_aggregation}. For further implementation details, please refer to the addressed code repository. Note that the $\text{binary\_step}(\psi,\; threshold=0.5)$ just converts a belief about the adjacency matrix into an standard adjacency matrix using a binary step function. The more complex of undefined functions is $\text{get\_path\_beliefs}$, that retrieves the beliefs for all possible paths between the intervened covariate to the disputed edge within the DAG, based on the most updated global knowledge.
\begin{algorithm}[ht]
\caption{Proximity-based federated aggregation method from interventional data.}
\label{alg:federated_setup_aggregation}
\begin{algorithmic}

\STATE
\STATE \textbf{function} \text{proximity\_based\_aggregation}$(\psi, \; \{\psi^1, \ldots, \psi^k\})$
\STATE \quad $adj\_mat \gets \text{binary\_step}(\psi,\; threshold=0.5)$ 

\InlineComment{latest structure from previous round's $\psi$}
\STATE \quad \textbf{for} $X_i \to X_j$ \textbf{in} $\text{edges}(adj\_mat)$ \textbf{do} 

\InlineComment{calculate reliability scores}
\STATE \quad \quad \textbf{for} $k = 1$ \textbf{to} $K$ \textbf{do}
\STATE \quad \quad \quad \textbf{for} $X_{s}$ \textbf{in} $X_{C_k}^{\mathcal{I}}$ \textbf{do}
\STATE \quad \quad \quad \quad $\psi^k_{sp}, \ldots, \psi^k_{ij} \gets \text{get\_path\_beliefs}(X_s, X_i \rightarrow X_j)$ 

\InlineComment{retrieves beliefs for all paths: $X_s \to X_p \to ... \to X_i \to X_j$}
\STATE \quad \quad \quad \quad $r_{ij}^k \gets max(\psi^k_{sp} \cdot \ldots \cdot \psi^k_{ij} \cdot m_{s}^k, \; r_{ij}^k)$ 

\InlineComment{over paths like $X_s \to \ldots X_j \to X_i$}
\STATE \quad \quad \quad \textbf{end for}
\STATE \quad \quad \quad $\hat{r}_{ij}^k = e^{\beta r_{ij}^k} / {\sum_{k\in[K]}e^{\beta r_{ij}^k}}$ 

\InlineComment{reliability scores to probabilities}
\STATE \quad \quad \textbf{end for}

\STATE \quad \quad $\psi^*_{ij} = \sum_{k\in [K]}\hat{r}_{ij}^k\psi_{ij}^k$ \InlineComment{final aggregation}
\STATE \quad \textbf{end for}
\STATE \quad \textbf{return} $\psi^*$

\end{algorithmic}
\end{algorithm}
\subsection{Graph and dataset generation}
\label{appendixb:datageneration}

We adhere to the random Erdős–Rényi \citep{erdos2011evolution} model to generate experimental DAGs of size $d = 20$. Erdős–Rényi graphs, named after mathematicians Paul Erdős and Alfréd Rényi, are a type of random graph model that provides a simple and fundamental framework for studying random network structures. These graphs are characterized by their random and independent edge formation. Each edge is added with an independent pre-determined probability from the others, and is modeled by $G(d=20,\;p)$. Hence, each edge is included in the graph with probability $p$. The probability for generating a graph that has $d$ nodes and $n$ edges is $p^n(1-p)^{{\binom{n}{2}} - n}$.

Our experiments are conducted for $ER-n$ graphs where $n \in \{1, 2, 4, 6\}$. The value of $m = n.d$ (where $n$ is the parameter in $ER-n$) is equal to the expected number of edges. For example, as $d=20$ throughout the experiments, an $ER-2$ DAG has an expected number of $m = 40$ edges. Further, to determine the orientation of the edges, we assume the causal ordering of $(X_i, X_j)$ to be oriented $X_i \rightarrow Xj$ if $i < j$, otherwise $X_j \rightarrow X_i$.

Same as \citet{lippe2021efficient} and \citet{ke2020dependency}, we employ categorical variables of 10 categories each. The process utilizes randomly initialized neural networks – MLPs with two layers with categorical inputs – to model the ground-truth conditional distributions. For instance, when a variable $X_i$ has $M$ parents, $M$ embedding vectors are stacked and passed as an input. The hidden size of the layers is 48, and their architecture uses a LeakyReLU activation functions between layers and a softmax activation to find a distribution for the output. More information about the process could be found in the appendix of \citet{lippe2021efficient}. 

\subsection{Baselines implementation}
\label{appendixb:implementation}
We use standard GitHub implementations of baselines in all cases. Aside from the size of the observational and interventional data required for our experiments, we keep other parameters of these models the same as their default values. Causal discovery toolbox~\citep{kalainathan2020causal} is the only third-party library that we use, others are all developed by the original implementations of the corresponding authors. \Cref{tab:baseline_implementation} presents a list of open source implementations used in our experiments with other baselines in the literature.

\begin{table}[h]
\centering
\caption{Baselines used in this work and the corresponding code repositories}
\label{tab:baseline_implementation}
\begin{tabular}{|l|l|}
\hline
\textbf{Method} & \textbf{Code} \\
\hline
ENCO & \url{https://github.com/phlippe/ENCO} \\
IGSP \& GIES & \url{https://github.com/FenTechSolutions/CausalDiscoveryToolbox} \\
SDI & \url{https://github.com/nke001/causal_learning_unknown_interventions} \\
NOTEARS-ADMM & \url{https://github.com/ignavierng/notears-admm} \\
FedDAG & \url{https://github.com/ErdunGAO/FedDAG} \\
\hline
\end{tabular}
\end{table}

\subsection{Hyperparameters}
\label{appendixb:hyperparams}
As appears in our methodology, there are a number of hyperparameters in our work that need further tuning. We mostly employ grid search for hyper parameter optimization. Note that for the initial mass, the value itself is not of critical importance. The importance lies in the difference between the injected mass for different clients when compared to each other. Therefore, we suggest setting the initial mass based on the size of interventional data for each client, and normalize this number across all clients between 0 and 1. We have tried setting the initial mass according to both interventional and observational data as well, and it can work better in some setups, especially if the size of observational data is critically low. Otherwise, the effect of choosing the mass based on observational samples becomes insignificant.

\begin{table}[h]
\centering
\caption{Hyperparameters and Their Effects}
\label{tab:hyperparams}
\begin{tabular}{|l|l|l|}
\hline
\textbf{Hyperparameter} & \textbf{Description} & \textbf{Plausible value(s)} \\
\hline
$\lambda_{prior}$ & Controlling the effect of prior belief on LCDMs local optimization& 0.01 - 0.1 \\
$m^k_{i}$ & The initial mass of clients for the aggreagation. & $\propto \frac{D_{\mathcal{I}^k}}{D_{\mathcal{I}}}, 0\leq m^k_{i}\leq1$ \\

$\beta$ & The softmax parameter for calculation of reliability scores. & 0.15 \\

\hline
\end{tabular}
\end{table}

\end{document}